\RequirePackage{fix-cm}
\documentclass{svjour3}
\usepackage{amsmath}
\usepackage{booktabs}
\usepackage{subfig}
\usepackage{graphicx}
\usepackage{multirow}
\usepackage[T1]{fontenc}
\usepackage{soul}
\usepackage{cancel} 
\usepackage{color}
\usepackage{url,float}
\newcommand\red[1]{\textcolor{red}{#1}}

\DeclareGraphicsExtensions{.pdf,.png,.jpg,.eps,.bmp}

\title{Diagnosis Support of Sickle Cell Anemia by Classifying Red Blood Cell Shape in Peripheral Blood Images}
\titlerunning{Erythrocyte Shape Analysis}

\author{Wilkie Delgado-Font \and Miriela Escobedo-Nicot \and
Manuel Gonz\'alez-Hidalgo \and Silena  Herold-Garcia \and 
Antoni Jaume-i-Cap\'o \and Arnau Mir} 

\institute{Wilkie Delgado-Font \at Facultad de Ciencias Naturales y Exactas, Universidad de Oriente, Santiago de Cuba, Cuba \and
Miriela Escobedo-Nicot \at Facultad de Ciencias Naturales y Exactas, Universidad de Oriente, Santiago de Cuba, Cuba \and
Manuel Gonz\'alez-Hidalgo \at Balearic Islands Health Research Institute (IdISBa), Soft Computing, Image Processing and Aggregation (SCOPIA) Research Group, Department of Mathematics and Computer Science, Universitat de les Illes Balears, Palma, Spain \and 
Silena  Herold-Garcia \at Facultad de Ciencias Naturales y Exactas, Universidad de Oriente, Santiago de Cuba, Cuba \and
Antoni Jaume-i-Cap\'o  \at Research Institute of Health Sciences (IUNICS), Computer Graphics and Vision and AI Group (UGiVIA), Department of Mathematics and Computer Science, Universitat de les Illes Balears, Palma, Spain,\\\email{antoni.jaume@uib.es} \and
Arnau Mir \at Balearic Islands Health Research Institute (IdISBa), Computational biology and bioinformatics (BIOCOM) Research Group, Department of Mathematics and Computer Science, Universitat de les Illes Balears, Palma, Spain}

\begin{document}

\maketitle

\begin{abstract}

\noindent{\sl Background and Objective}: Red blood cell (RBC) deformation is the consequence of several diseases, including sickle cell anemia, which causes recurring episodes of pain and severe pronounced anemia. Monitoring patients with these diseases involves the observation of peripheral blood samples under a microscope, a time-consuming procedure. Moreover, a specialist is required to perform this technique, and owing to the subjective nature of the observation of isolated RBCs, the error rate is high. 
\vskip 0.5 cm 
\noindent{\sl Methods}: In this paper, we propose an automated method for differentially enumerating RBCs that uses peripheral blood smear image analysis. In this method, the objects of interest in the image are segmented using a Chan-Vese active contour model. An analysis is then performed to classify the RBCs, also called erythrocytes, as normal or elongated or having other deformations, using the basic shape analysis descriptors: circular shape factor (CSF) and elliptical shape factor (ESF). To analyze cells that become partially occluded in a cluster during sample preparation, an elliptical adjustment is performed to allow the analysis of erythrocytes with discoidal and elongated shapes. The images of patient blood samples used in the study were acquired by a clinical laboratory specialist in the Special Hematology Department of the ``Dr. Juan Bruno Zayas'' General Hospital in Santiago de Cuba. 
\vskip 0.5 cm 
\noindent{\sl Results}: A comparison of the results obtained by the proposed method in our experiments with those obtained by some state-of-the-art methods showed that the proposed method is superior for the diagnosis of sickle cell anemia. This superiority is achieved for evidenced by the obtained F-measure value (0.97 for normal cells and 0.95 for elongated ones) and several overall multiclass performance measures.
\vskip 0.5 cm 
\noindent{\sl Conclusions}: The results achieved by the proposed method are suitable for the purpose of clinical treatment and diagnostic support of sickle cell anemia.
\end{abstract}


\section{Introduction}
\label{intro}
Sickle cell anemia is a genetic disease characterized by a 
	modification in the shape of red blood cells (RBCs) from a smooth donut shape to 
	a crescent or half-moon shape. The misshapen cells lack plasticity and can 
	block small blood vessels, impairing blood flow, which leads to 
	shortened RBC survival and subsequently causes anemia. This condition has been defined as a health priority by the World Health Organization (WHO) \cite{WHO}.
	
	Without appropriate treatment, low blood oxygen levels and blood vessel blockages 
	cause continuing episodes of pain, severe pronounced 
	anemia, severe bacterial infections, necrosis (tissue death), recurrent 
	infections, a drastic decline in the patient's quality of life, clinical 
	symptoms of thrombosis in internal organs that can lead to disorders of the 
	central nervous system, episodes of neurophysiological disorders, and even 
	the patient's death. 
	
	Some facts about sickle cell disease provided by  \cite{WHO} \red{in 2011} are as follows:
	\begin{itemize}
		\item Approximately 5\% of the world's population carries trait genes for hemoglobin disorders, mainly sickle cell disease and thalassemia.
		\item The percentage of people who carry these genes is as high as 25\% in some regions.
		\item Sickle cell disease predominates in Africa.

		\item Over 300,000 babies with severe hemoglobin disorders are born each year.

	\end{itemize}
The monitoring of these patients is therefore essential. It includes the 
	observation of peripheral blood samples under a microscope, a time-consuming procedure. In addition, this procedure must be performed by a specialist, and  because of the 
	subjective nature of the observation, its error rate is high. These problems are aggravated when the number of patients is large. 
	A method to evaluate the clinical status of patients is to enumerate the types 
	of RBCs based on their morphology. They can be normal (discocyte) or 
	deformed elongated (sickle cells) or have other deformations. This quantitative 
	assessment presents several problems, ranging from experts'differences of opinion to the difficulty of establishing a standard 
	evaluation.

In the morphological analysis of digital image structures for  subsequent application in appropriate classification algorithms, the process of extracting the characteristics of interest from the analyzed edge is critical. To analyze the deformation in the erythrocytes that are present in peripheral blood samples, several methods have been designed, each with its own characteristics, and their performance results have been reported. We now describe several of these methods.

\subsection{State of the art}
 An automated analysis method for the rapid classification of large numbers of RBCs from individual specimens was presented in \cite{taherisadr2013new}. The method is based on digital image processing. Several features related to the RBCs'shape, internal central pallor configuration, circularity, and elongation are extracted and, with the help of a decision  process, the cells are classified into 12 categories. This method differs from others in that datasets are not required to learn the model and thus allow its use. Moreover, in the study in~\cite{taherisadr2013new}, cluster cells were not analyzed.
	
Morphological studies of sickle erythrocytes (SS cells) using a computer-assisted image analysis system were described in \cite{asakura1996percentage}. The system determines the percentage of SS cells having various morphologies by measuring the area, perimeter, and short-axis/long-axis ratio of each cell. The circular shape factor (CSF) and elliptical shape factor (ESF) are determined. The obtained results demonstrated marked temperature-dependent differences in RBC morphology and showed the percentages of reversible sickle cells (RSCs) and irreversible sickle cells (ISCs). Cluster cells were not considered in~\cite{asakura1996percentage}. The author of \cite{frejlichowski2010pre} presented an approach for classifying the deformations of erythrocyte shapes for diagnostic purposes. In this approach, the contour shapes of binary images are represented by three shape descriptors based on polar methods. They are matched with basic elements to identify those with the greatest similarity. \red{Many types of biomedical images share classification methods and types of features extracted such as shape, color and texture features, see for instance \cite{Ma2017effective,Oliveira2018computational}}.

In the studies presented in \cite{hirimutugoda2010image,purwar2011automated}, the detection of malaria infections in peripheral blood samples was  investigated. The image-based method proposed in \cite{purwar2011automated} constitutes an unsupervised screening method for diagnosing malaria by using thin smear analysis. The method was designed to automate the process of enumerating the cells and identifying the parasites. The goal was to distinguish between samples that were positive and negative for malaria using thin smear blood slide images. For this reason, the morphological aspect of the cell was not considered. In this method, the segmentation of the cell is achieved by applying the Chan-Vese method. Because of the circular shape of the cell in the image presented in this paper, it can be fitted using a Hough Transform.  In \cite{hirimutugoda2010image}, the authors described the automatic detection of abnormal erythrocytes using an artificial neural network in order to locate malarial parasites and detect thalassemia in blood sample images. Two back-propagation artificial neural network models (of three and four layers) are  used together with image analysis techniques to evaluate the classification accuracy for recognizing in a medical image the patterns associated with the morphological features of erythrocytes in the blood. The principal disadvantage of this method lies in the need to prepare a large dataset for training and testing. Frequently used image processing operations are performed on the digital images, including image enhancement, edge erosion, and color and size normalization. 

Research was conducted on samples stained with peripheral blood in the study presented in \cite{sabino2004texture,eom2006leukocyte,chen2016automatic} to investigate WBCs (also called leukocytes). In \cite{sabino2004texture}, the authors proposed a segmentation method of leukocytes in blood smear images. This method is based on a region-based active contour model that drives the initial contours toward the boundary of a leukocyte, thus avoiding initialization and local minima problems. This paper focused only on measurements of the accuracy of the segmentation method. In \cite{eom2006leukocyte}, a framework developed for leukocyte recognition based on the interactive color
segmentation of blood smears, feature extraction, and the statistical classification of feature vectors was presented. In the framework, four textural attributes, energy, entropy, inertia, and local homogeneity, are calculated based on gray-level co-occurrence matrices (GLCM). In the study, these features were tested to ascertain their effectiveness for leukocyte recognition.  Data mining algorithms were implemented to estimate suitable scales. Feature selection methods were also applied to determine the most discriminative attributes for describing the cellular patterns. The experimental results showed that texture parameters are suitable for differentiating between the five types of normal leukocytes and those showing chronic lymphocytic leukemia.
    
	An automatic system was proposed to recognize the position and number of erythrocyte nuclei and the segments of complex erythrocyte nuclei, and to count the number of nuclei by using the Otsu method in \cite{chen2016automatic}. According to the number of observed nuclei, normal erythrocytes can be distinguished from those showing megaloblastic anemia. In this method, observation of the RGB color space is used to automatically segment and separate clustered nuclei by using distance transforms and drawn-circle techniques. In the study in~\cite{chen2016automatic}, a field of the samples cells was taken into account, but cluster cells were not considered.

\red{Recall that segmentation algorithms, in addition to being used to analyze images of  blood samples, have also been used in the analysis of other types of biomedical images.  In  \cite{Ma2009review}, \cite{Ma2010review}, and \cite{Ferreira2014segmentation} we can find several interesting review on segmentation algorithms for different types of biomedical images. Moreover, a review on physical based segmentation algorithm can be found in \cite{gonccalves2009segmentation}. }

In \cite{ritter2007segmentation} and \cite{yao2007blood}, erythrocyte segmentation methods were proposed and the assessment of their effectiveness was presented. In \cite{ritter2007segmentation},  a fully automated algorithm for locating every object (cell or cell group) in an image taken from a canine peripheral blood smear slide was presented. This method separates cells that just touch, and operates on normal, deformed, and joined RBCs, as well as on WBCs and RBC fragments. In the study, cluster cells were not taken into account. In contrast, in \cite{yao2007blood} an alternative RBC segmentation algorithm based on statistical shape analysis was proposed. The authors'goal was to design a method that automatically recognizes the blood cells, including overlapped cells, by means of correct segmentation. Their study was focused only on segmentation and did not include a morphological analysis.	
	 
	For the classification of individual cells, two novel approaches were presented in \cite{gual2015erythrocyte,gual2015geometric}. In  \cite{gual2015erythrocyte}, very efficient results were reported for the supervised classification of erythrocytes by using functions based on integral geometry for contour features, some of which were previously proposed in \cite{ren1994topics,gual2013shape}. In the study in \cite{gual2015geometric}, contours were considered as elements in the shape space and the classification was obtained using templates of a circle and an ellipse.

In \cite{fernandez2013estudio}, a learning matrix was described that takes into account the textures of the test images by applying a semi-supervised learning process. The obtained learning matrix  $ M_a $ contains textures of different objects in the images, which are classified as belonging and not-belonging to erythrocytes. To obtain a new image from the contours of the detected regions, segmentation algorithms based on active contours are initialized. The algorithms evolve to determine the edges of the objects. Finally, a morphological and classification analysis of the erythrocytes is conducted, which obtains good results in general. An assessment of the results of a case study involving four patients (two healthy and two with sickle cell disease) was presented. However, the method has a disadvantage: it cannot be used to classify cell clusters. In addition, it requires a training process, which adds considerably to the amount of time required to execute the entire method.

Additional automatic counters of RBCs based on erythrocyte shape descriptors were presented in the literature~\cite{acharya2018identification,frejlichowski2010preprocessing}. A recently developed simple method based on the CSF feature was presented in ~\cite{acharya2018identification}. The method presented in \cite{frejlichowski2010preprocessing} is based on a set of shape descriptors using polar transform and 2D Fourier transform. As we compared these methods with the method proposed in this paper, we describe them in detail in the results sections.
 
In the study in \cite{gonzalez2015red}, ellipse adjustments obtained by a new algorithm for detecting notable points were used. A set of constraints was applied that allowed the elimination of significant image preprocessing steps proposed in previous papers. Three types of images were obtained. For each type of image, a high efficiency value was achieved. The method was not applied to the entire field of view of the sample, but only to independent regions of clustered cells.

\subsection{Objectives}
Because of the shortcomings of the methods mentioned above, in this paper we present a new method to obtain erythrocyte classification in normal, elongated, and cells with other deformities using peripheral blood smear sample images. The method uses a Chan-Vese active contour model \cite{chan2000active} to segment objects in images. The Chan-Vese model was selected because it achieves good performance in the image segmentation task as a result of its ability to obtain a wider range of convergence and handle topological changes in a natural manner, as the examples provided in this paper show. It provides excellent image segmentation. The obtained borders are analyzed to determine whether the detected objects are cells, cell clusters, or other objects in the image. For single cells, the elementary coefficients of circularity and ellipticity proposed in \cite{asakura1996percentage} are used. For cell clusters, an elliptical adjustment proposed in \cite{gonzalez2015red} is applied. The results are compared with those obtained in \cite{acharya2018identification,fernandez2013estudio,frejlichowski2010pre,gual2015erythrocyte}.
	
\subsection{Structure of the paper}
The rest of this paper is organized as follows. In Section \ref{section2}, we describe the proposed method for erythrocyte classification, which uses the Chan-Vese model for image segmentation, as well as an ellipse adjustment for cluster analysis. In addition, in this section, we  explain the measures used to classify the erythrocytes and the method used to classify cells in a cluster using ellipse adjustment.  Section 3 is devoted to the experimental environment \red{and to set the parameters values of each step of the proposed method}. A description of the database used in this study is included, together with a description of three different experiments designed to validate our method using this public database. An analysis of the results obtained from the experiments is presented and discussed in Section \ref{section4}. Finally, we present some concluding remarks.

\vskip 0.5 cm

\section{Materials and Methods}\label{section2}

%\subsection{Outline of the method}

Our method consists of the following steps:
	\begin{enumerate}\label{steps}
		%\item Sample preparation and image acquisition.
		\item Segmentation using the Chan-Vese model (active contour).
		\item Evaluation of cell areas. Separation of isolated cells from cell clusters. 
		\item Classification of cells in each cluster using ellipse adjustment and the \textit{CSF}.
		\item Implementation of elemental shape descriptors: classification of isolated cells using \textit{CSF} and  \textit{ESF.} 
		\item Cell count.
	\end{enumerate} 

\begin{figure}[h!]
	\centering
	\includegraphics[width=0.8\linewidth]{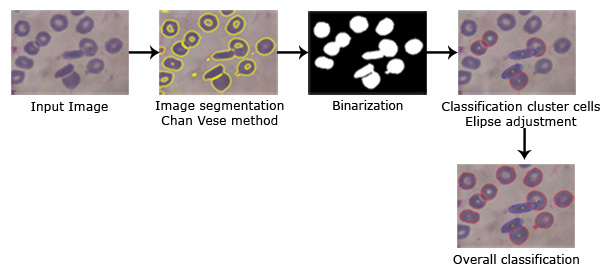}
	\caption{Workflow of the process.}
	\label{fig:flow}
	\end{figure}

Briefly, the design of the proposed algorithm is as follows. The Chan-Vese method is applied to RBC images without any preprocessing to obtain segmentation. The result is a binarized image, from which small objects that can interfere with the classification have been removed. \red{In order to do that, we removed all connected components (objects) that have less than $P$ pixels by applying an area opening. We set $P=80$}. When these \red{small} objects have been removed, noise and the remaining information are considered regions belonging to the cells. Next, the area of each ``region'' is computed and the regions larger than 1.4 times the average area of all the regions are considered clusters. These clusters are processed using an ellipse adjustment method that is explained below. The regions that comply with the previous restriction are classified using ESFs and CSFs.  Figure \ref{fig:flow} shows the most important steps in the process workflow. A more detailed description of each step of the algorithm is provided below. 

\subsection{Image Segmentation: Chan-Vese Method}
	
		In this study, we used the Chan-Vese method \cite{chan2001active} to segment the blood smear images. This is an active contour model based on Mumford-Shah segmentation techniques \cite{mumford1989optimal} and the level set method \cite{osher1988fronts}. We used the implementation provided in \cite{chan2000active}. The Chan-Vese model is a curve evolution implementation of a piece-wise constant case of the Mumford-Shah model \cite{mumford1989optimal}. The Mumford-Shah model is an energy-based method that uses an energy functional, which was introduced by Mumford and Shah in \cite{mumford1989optimal}. Let $ f $ denote the given grayscale image on a domain $ \Omega $ to be segmented. The Mumford-Shah model approximates the image $ f $ by a piece-wise smooth function $ u $ as the solution of the minimization problem:
		\begin{equation}\label{mumford}
		\arg\min_{u,C}   \mu\cdot \ell (C)+\lambda \cdot \int_{\Omega}(f(x)-u(x))^2
		dx + \int_{\Omega \setminus C}|\nabla u(x)|^2 dx,
		\end{equation}
where $C\subset \Omega$ is set of curves, $\ell (C)$ is its length, and  $u$ can be discontinuous on $C$. The first term guarantees that $ C $ is as short as possible and its regularity. The second term guarantees that the piece-wise approximation $ u $ is as similar as possible to $f$. The third term guarantees that $ u $ is smooth over $\Omega \setminus C$. Mumford-Shah  suggests in \cite{mumford1989optimal} to take the set $ C $ as the segmentation of image~$f$.
		
The Chan-Vese model differs from the Mumford-Shah model in two main cues, namely: a term penalizing the enclosed area by $C$ is added and the function $u$ is allowed to have only two values $c_1$ and $c_2$, that is:
\[
u(x)=\begin{cases}
c_1, & \mbox{where $x$ is inside $C$, } \\
c_2, & \mbox{where $x$ is outside $C$,}
\end{cases}
\]
where $ C $ is the boundary of the enclosed set. The Chan-Vese model approximates the image $f$ by a function $ u $ given by equation~(\ref{eqchan}) as the best approximation obtained minimizing the functional
		\begin{equation}\label{eqchan}
		\begin{split}
		\arg\min_{u,C}  \, & \mu \,\ell (C)+v \mbox{Area}(\mbox{inside}(C))\\
		&+\lambda_1\int_{\mbox{inside}(C)} |f(x)-c_1(x)|^2dx \\
		&+\lambda_2\int_{\mbox{outside}(C)} |f(x)-c_2(x)|^2 dx,
		\end{split}
		\end{equation}	
The first term has the same goal that in the Mumford-Shah functional. The second term penalizes the size of the area enclosed by $ C $, controlling its size using~$v$. The two last terms penalize the difference between the approximation $ u $ and the original image $ f $. When the local minimum for this problem has been determined, the segmentation of the image $f$ is obtained as the best two-phase piece-wise constant approximation~$u$.
		
		Solve the problem done by equation~(\ref{eqchan}) requires its minimization over the set of possible boundaries. This is achieved by applying the level set technique introduced by Osher and Sethian \cite{osher1988fronts}. So, the boundary $C$ is represented by a level set function $ \phi $: 
		\begin{equation}\label{key}
			C=\{x \in \Omega\, :\, \phi(x)=0\}.
		\end{equation}
Using the previous expression, we can write the sets ``inside $C$''  and ``outside $C$'' as:
\[
\mbox{inside}(C) = \{x \in \Omega\, :\, \phi(x) > 0\}\,\, \mbox{and }\, \mbox{outside}(C) = \{x \in \Omega\, :\, \phi(x) < 0\}.
\]

Then, the Chan-Vese segmentation method can be rewritten using the Heaviside function $H(t)$ and the Dirac delta function $\delta(t)$, defined by: 
\begin{equation}\label{heavyside}
		H(t) =
		\left\{
		\begin{array}{llcc}
		1,  & \mbox{if } t \geq 0, \\
		
		0, & \mbox{if } t < 0,
		\end{array}
		\right.
		\hspace{0.3 cm}\delta(t)=\dfrac{d}{dt}H(t),\\
		\end{equation}
in terms of the level set function $ \phi $ as
		\begin{equation}\label{key1}
		\begin{split}
		\arg\min_{c_1,c_2,\phi} &  \mu \int_{\Omega}\delta(\phi(x))|\nabla\phi|dx + v\int_{\Omega}H(\phi(x))dx\\
		&+\lambda_1\int_{\Omega}|f(x)-c_1|^2H(\phi(x))dx\\
		&+\lambda_2\int_{\Omega}|f(x)-c_2|^2(1-H(\phi(x)))dx,
		\end{split}
		\end{equation}

Note that the term penalizing the area appears as the integral of $H( \phi $) and the first term is the length of~$C$. Here the $H(t)$ and $\delta(t)$ functions are used to divide the level set into the three formed parts: the part inside~$C$, the part outside~$C$, and the segmentation boundary~$C$, avoiding the numerical problems that appears using the level set function~$\phi$.

		The minimization problem is solved by alternately updating $c_1$, $c_2$, and $ \phi $. For a fixed $ \phi $, the optimal values of $c_1$ and $c_2$ are the region averages:
		\begin{equation}
		c_1(\phi)=\frac{\displaystyle  \int_{\Omega}u_0(x,y)H (\phi(x,y))dxdy}{\displaystyle   \int_{\Omega}H (\phi(x,y))dxdy},\quad 
		c_2(\phi)=\frac{\displaystyle \int_{\Omega}u_0(x,y)(1-H (\phi(x,y)))dxdy}{\displaystyle  \int_{\Omega}(1-H (\phi(x,y)))dxdy}.
		\end{equation}
	
	The advantages of the Chan-Vese method versus the traditional  level sets methods are reported in \cite{chan2000active,chan2001active}, and are also pointed out in \cite{maroulis2007variable,wang2010efficient} in an image segmentation environment. The principal advantage is that this method can handle the detection of smooth boundaries without using gradients and their interiors in an automatic way. This implies that the method can be applied to the original image without any preprocessing, even in presence of noise. Moreover the initial curve can be placed anywhere in the image. These are the main reasons why we have chosen the previously described method to segment the images of peripheral blood smear, in order to subsequently classify and count the RBCs.

\begin{figure}[!ht]	
\centering
\subfloat[][]{\includegraphics[width=.23\textwidth]{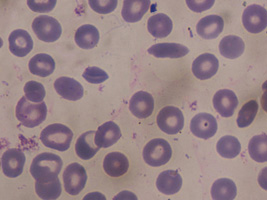}}\quad
			\subfloat[][]{\includegraphics[width=.23\textwidth]{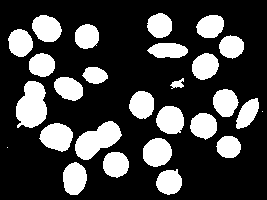}}\quad
			\subfloat[][]{\includegraphics[width=.23\textwidth]{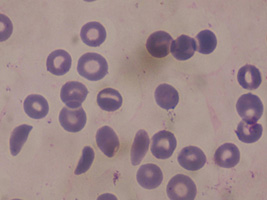}}\quad
			\subfloat[][]{\includegraphics[width=.23\textwidth]{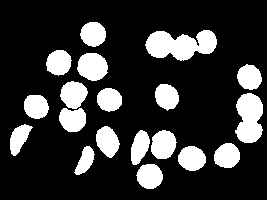}}
			 \caption{(a) Original input image. (b) Result of Chan-Vese segmentation for image~(a). (c)  Original input image. (d) Result of Chan-Vese segmentation for image~(c).}\label{fig:count_a}
\end{figure}

The output of this step is a binarized image, from which the areas corresponding to isolated cells or a cluster of cells, together with their contours, can easily be extracted. Before the extraction of cells or clusters of cells, the small objects that may interfere with the classification are removed. In addition, all detected areas that touch the border of the image are eliminated and are not taken into account in the following steps of the method. In general, it is very difficult to classify an object that is on the border of an image, only a part of which is visible. Some examples are displayed in Figure~\ref{fig:count_a}, where we show the original image and its segmentation obtained using the method previously described.

\subsection{Evaluation of cell areas. Separation of isolated cells from cell clusters}

After the Chan-Vese segmentation of an original peripheral blood smear image has been obtained, we proceed to extract the image areas that correspond to isolated cells and those that are groups of different cells. 

Let $\{ A_i \, :\, i=1,\ldots , n\}$ be the set of different  detected areas  for one image. By $|A_i|$ we denote the area of region $A_i$. We define $|A_m|$ as the mean area of all objects in the image, that is:
\[
|A_m| =\frac{\sum_{i=1}^n|A_i|}{n}.
\]

To classify the regions obtained as single cells or overlapped cells, we consider that overlapping cell regions follow the condition
	\begin{equation}\label{Region}
	|A_i|\geq |A_m|\cdot k,
	\end{equation}
where $k$ is a constant that we determine by systematic experimentation, the  value of which is set to $1.4$.

\subsection{Classification of cells in a cluster using ellipse adjustment}\label{elipse}
	
	When clusters that include elongated or elliptical objects appear in digital images of blood samples, the classification of erythrocytes can be problematic. In the first step,  the cells and clusters of overlapping cells are detected, for which a segmentation method is applied. After obtaining cluster contours, algorithms are applied for the detection of concave points to identify the objects in each cluster.
	
	In this study, we used the method proposed in \cite{gonzalez2015red}, in which ellipses to detect objects with a circular or elongated shape in a cluster are used, as well as new constraints to determine valid objects. To compute the points of interest, the changes in curvature, both vertical and horizontal, in the contour that represents the cluster are considered. This method also facilitates the efficient detection of concave points in the contour and almost completely avoids the points that correspond to contour noise concavities or details of the contour. This algorithm allows  the concave points of interest to be obtained efficiently, which affects the subsequent performance of the algorithms used to detect objects in a cluster. 
	
	The circumference adjustment method is not suitable for clusters that include elongated or elliptical objects. Given two adjacent concave points, an ellipse is obtained by a least squares adjustment of the arc formed by  $ n $ points of the contour between these two previous points. This least-squares adjustment requires a parametric model that relates the data obtained to the predicted data using one or more coefficients. The result of the adjustment process is an estimate of the coefficients of the model. We now explain the method in detail. Given  $n$ points, $ {(x_i,y_i)}_{i=1}^{n} $, the method attempts to find the best ellipse, the general equation of which is
	\begin{equation}\label{ellipseEq}
	ax^2+bxy+cy^2+dx+ey+f=0,
	\end{equation}
where   $4ac-b^2>0$. The least squares adjustment method finds the coefficients $ a $, $ b $, $ c $, $ d $, $ e $, and $ f $ that minimize the adjustment error of the $ n $ points:
	\begin{equation}\label{ellipseSys}
	S=\sum_{i=1}^{n}((ax_i^2+bx_iy_i+cy_i^2+dx_i+ey_i)+f)^2.
	\end{equation}
We can write the previous expression using matrices and vectors:
	\begin{equation}\label{matrixREP}
	S=(XA+F)^\top (XA+F),
	\end{equation}
	where $ A=(a,b,c,d,e,f)^\top $ is the parameter vector, $ F =(f,\ldots,f)^\top$ ($n$ components), and $ X  $ is a matrix $n\times 5$, the $i$-th row of which is $(x_i^2,x_iy_i,y_i^2,x_i,y_i)$.  
By developing the previous expression, we obtain
	\begin{equation}\label{key2}
	S=A^\top X^\top XA+2F^\top XA +  f^\top f.
	\end{equation}
	
  To minimize $S$ in order to determine the coefficients~$A$, we must derive $S$ with respect to each parameter and set the result to zero: 
	\begin{equation*}\label{key3}
	\dfrac{\partial S}{\partial A}=2X^\top XA+2F^\top X=0.
	\end{equation*}
Therefore, to compute the coefficients~$A$ we must solve the following linear system of equations.
    \begin{equation*}
    (2X^\top X) A =-2F^\top X.
    \end{equation*}

When the coefficients of the ellipse have been obtained, in the next step  the canonical form of the ellipse is determined. The expressions of the minor axis $\alpha_1$, the major axis $\alpha_2$, and the center of the ellipse $(x_0,y_0)$ can be calculated using the following well known equations.
	\begin{equation}\label{key44}
	\alpha_1=\sqrt{\left|\dfrac{f_1}{a_1}\right|}, \alpha_2=\sqrt{\left|\dfrac{f_1}{c_1}\right|},x_0=-\dfrac{d_1}{2a_1},y_0=-\dfrac{e_1}{2c_1},
	\end{equation}
where
	\begin{equation*}
	\begin{split}
	a_1=a(\cos\Phi)^2-b\cos\Phi\sin\Phi+c(\sin\Phi)^2,\\
	c_1=a(\sin\Phi)^2+b\cos \Phi\sin\Phi+c(\cos\Phi)^2,\\
	d_1=d\cos\Phi-e\sin\Phi,e_1=d\sin\Phi+e\cos\Phi,\\
	f_1=-f+\dfrac{(d_1)^2}{4a_1}+\dfrac{(e_1)^2}{4c_1},\Phi=\dfrac{1}{2}\arctan\left(\dfrac{b}{c-a}\right).
	\end{split}
	\end{equation*}

Therefore, the canonical equation of the ellipse is
    \[
    \dfrac{(x-x_0)^2}{\alpha_1^2} + \dfrac{(y-y_0)^2}{\alpha_2^2} =1.
    \]
The ellipse is valid because it has been adjusted in relation to the concave points. The criteria include the relationship between the areas of the tight-fitting object and the detected ellipse, as well as the percentage of object occlusion. The following criteria (see \cite{gonzalez2015red}) are checked when a cluster $ S $ and the ellipse $ E_i $ found for an arc between two adjacent concave points $ i  $ and $ i + 1 $ are analyzed.
	
	\begin{itemize}
		\item $ \dfrac{area(D_i)}{area(E_i)}>\tau_{e1} $, where $ D_i=S\cap E_i $ and $ \tau_{e1} $ is a threshold that controls the relationship between the areas of the tight-fitting object and the detected ellipse.  
		\item $ \dfrac{area(E_i)}{S_{E_k}} > \tau_{e2} $, where $ S_{E_k} $ is the cluster of all ellipses that differ from $ i $, $ S_{E_k} = \{\cup_{k=1}^{n}E_k |k \neq i\}$, where $ n $ is equal to the number of ellipses detected and $ \tau_{e2} $ is a threshold that controls the occlusion percentage of objects.
	\end{itemize}
For a detailed account about what guarantees each one of these thresholds see \cite{gonzalez2015red}.
\vskip 0.5 cm 

	If the arc between the pairs of concave adjacent points $ i $ and $ i + 1 $ does not provide sufficient information to allow the ellipses to be adjusted or if the adjustment is incorrect (i.e., it does not meet the criteria described above), an additional arc is added. This arc is circumscribed by the pairs of concave adjacent points $ j $ and $ j + 1 $, where $ j\neq i $, $ j \neq i + 1 $, and $ j + 1 \neq i $. This information is used to perform an additional adjustment and for subsequent validation of the new ellipse.
    
\begin{figure}[ht]
\centering
\begin{tabular}{cccc}
 \subfloat[]{\includegraphics[width=0.25\textwidth]{HCSS02C2.jpg}} &  \subfloat[]{\includegraphics[width=0.25\textwidth]{HCSS02C2_ChV.png}} & \begin{tabular}{cc}
 \subfloat[]{\includegraphics[width=0.055\textwidth]{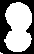}} \\
  \subfloat[]{\includegraphics[width=0.15\textwidth]{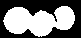}}
\end{tabular} & 
\begin{tabular}{cc}
 \subfloat[]{\includegraphics[width=0.06\textwidth]{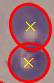}} \\
  \subfloat[]{\includegraphics[width=0.15\textwidth]{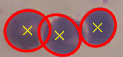}}
\end{tabular}
\end{tabular}
\caption[]{(a) Original image. (b) Chan-Vese segmentation. (c) and (d) Cell cluster detected in (b). (e) and (f) Objects identifies by ellipse adjustment to (c) and (d).}
\label{fig:eadjust}
\end{figure}

When the ellipse adjustment has been realized, the number of objects in the cluster corresponds to the number of recognized elliptical objects. Then, the elliptical coefficient (the ESF) (see below for the description) is used to determine the class of cells to which each detected object corresponds. With the aim of achieving better visualization, an ellipse or circumference is painted on the original image, centered on the centroid of the detected object. In Figure \ref{fig:eadjust}, we can see the results of this step for the clusters detected in one of the images in our dataset.

\subsection{Elemental shape descriptors. Classification of isolated cells using circular shape factor and elliptical shape factor }\label{ESFyCSF}
	
Among the most basic descriptors for shape analysis are the circularity coefficient and the elliptical coefficient \cite{asakura1996percentage}, which are based on the geometric features of the objects.
	
	\textbf{Circularity coefficient (the \textit{CSF})}\\
	The number of geometric parameters used in the comparison of observed objects from different distances (not dependent on object size) is minimal for circular objects, but is sensitive to noise, because noisy edges have a very long perimeter as compared with smooth edges. The number is defined as
	\begin{equation}\label{eq1}
	CSF=\dfrac{4\pi A}{p^2},
	\end{equation}
where $A$ is the area and $p$ the perimeter. 

	\textbf{Elliptical coefficient (the \textit{ESF})}\\
	The elliptical coefficient expresses the elongation of the object. It takes into account the ratio between the largest and smallest abscissas of the object, which are equal to 1 if the object 
	is circular.
	\begin{equation}\label{eq2}
	EFS=\dfrac{ME}{MA},
	\end{equation}
	where $ ME $ and $ MA $ are the minor abscissa and the major abscissa of the object, respectively. 
	
	To determine whether the segmented cells are deformed, the values reported in \cite{fernandez2013estudio} were used as the coefficients of compactness and ellipticity. A verification is performed to validate the values'effectiveness. The considered criteria are defined as
	
	\begin{itemize}
		\item If $ EFS < 0.5, $ the cell is considered a deformed elongated cell.
		\item If $ EFS > 0.5 $ and $ CSF < 0.8 $, the cell is considered a deformed slightly elongated cell, which 
		includes starry and leaf.
		\item If $ EFS > 0.5 $ and $ CSF > 0.8 $, the cell is considered a normal discoid.
	\end{itemize}

\subsubsection{Counting the cells}    

\begin{figure}[!ht]
\centering
\subfloat[][]{\includegraphics[width=.23\textwidth]{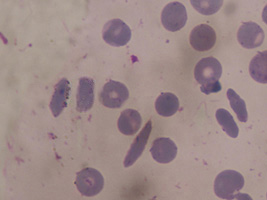}}\quad
			\subfloat[][]{\includegraphics[width=.23\textwidth]{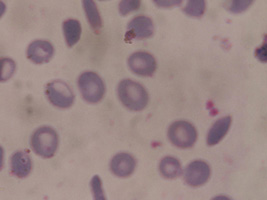}}\quad
			\subfloat[][]{\includegraphics[width=.23\textwidth]{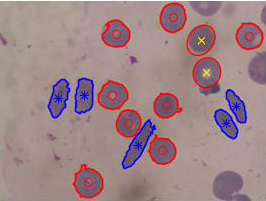}}\quad
			\subfloat[][]{\includegraphics[width=.23\textwidth]{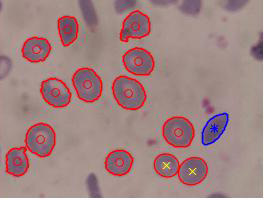}}
			\caption{(a) and (b) Original input images. (c) and (d) Classification results using the proposed algorithm for (a) and (b), respectively. \red{Cells detected as belonging to the normal class are marked in red and cells belonging to the elongated class are marked in blue.}}
			\label{fig:count}
\end{figure}

Finally, when all the areas have been classified, we count the number of areas that have been classified into each of the classes considered: normal, elongated, and other deformations. The output of the algorithm is the confusion matrix that allows  the performance of the algorithm to be measured. Each row of the matrix represents the number of erythrocytes belonging to a real class and each column represents the number of erythrocytes belonging to a predicted class. In addition, the algorithm provides an image in which the edges of the different detected shapes overlap the original image; each class is shaded with a different color and a mark appears in the centroid in each of the detected cells. The cells belonging to the normal class are marked with a circle, those belonging to the elongated class with a $\ast$, and the remaining cells with a $+$. An ``X'' is also drawn in the center of each cell detected in a cell cluster. Examples of the results of the process are shown in Figure \ref{fig:count}.

\begin{figure}[h!]
	\centering
	\includegraphics[width=0.7\linewidth, height=.35\textheight]{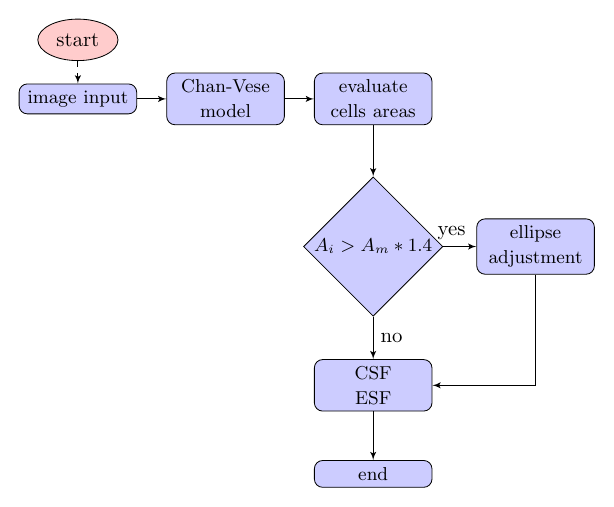}
	\caption{Process workflow.}
	\label{fig:flow1}
	\end{figure}

In Figure \ref{fig:flow1}, the block diagram of the algorithm describing the RBC three-class classifier is shown. 
    
\section{Description of the Experiments}\label{section3}

%In this paper, we \red{have} used two different experiments to validate our method. The \red{obtained} results \textst{obtained} were compared with \cite{fernandez2013estudio} and \cite{gonzalez2015red}. The first experiment was designed to validate the effectiveness of the cell detection, and the second experiment was designed to validate the classification step and check its performance. \red{Next,} \textst{Below} \red{a} \textst{are} brief description\textst{s} of each experiment is shown.
In this study, we conducted three different experiments to validate our method over a public dataset. The first experiment was designed to validate the effectiveness of the cell detection.  Then, the second experiment was designed to validate the classification step and examine its performance. Finally, in the third experiment we compared the performance of our proposed method with that of state-of-the-art methods for automatically counting RBCs based on morphology. Next, a brief description of the dataset and each experiment is provided.

\subsection{Dataset \red{and Experimental Framework}}
The dataset used in this study is available at \url{http://erythrocytesidb.uib.es/}. The images used were obtained with the assistance of a first-grade specialist at the Clinical Laboratory for Special Hematology at the ``Dr. Juan Bruno Zayas'' Hospital General in Santiago de Cuba, Cuba. This specialist took and prepared samples of patients with sickle cell anemia and then analyzed the images manually to classify the cells as normal, elongated, or with other deformations. The specialist's criteria were used as an expert approach for validating the results of the methods applied to the classification of the cells present in the images. 

Samples were obtained from voluntary donors. The donor's thumb was pricked with a lancet and a drop of blood was collected on a sheet. The blood was spread with a coverslip at an angle of $ 45^{\mathrm{o}} $ with respect to the sheet, allowed to dry, and then fixed with a May-Gr\"unwald methanol solution. May-Gr\"unwald-Giemsa staining is a frequently used method for blood smears. 
    
The images were acquired in the Department of Special Hematology of the Clinical Laboratory at ``Dr. Juan Bruno Zayas'' Hospital General, using a Leica microscope (100$ \times $) and a Kodak EasyShare V803 camera (Kodak Retinar Aspheric All Glass Lens of 36--108 mm AF 3$ \times $ optical). Forty-five images of size 500 x 375 pixels with a resolution of 480 dpi (dots per pixel) were obtained. 

\vskip 0.5 cm 

\red{
We should point out that all the experiments have been carried out on an Intel(R) Core(TM) i7-3632QM CPU @ 2.20GHz PC with   4.00GB of RAM. To give an idea of the computational cost of our algorithm, we will give the execution times of the image (c) of Figure~\ref{fig:count_a} as a model of the algorithm steps in addition to the average execution times of all images in the database:
}

\red{
\begin{itemize}
\item	Chan-Vese segmentation (image (d) of Figure~\ref{fig:count_a}): 1.421 s. Average time of all images: 4.59 s.
\item	Detection and classification of individual cells: 0.508 s. Average time of all images: 0.66 s.
\item	Overlapping and classification of all the cells: 6.185 s. Average time of all images: 3.14 s.
\end{itemize}
}

\vskip 0.5 cm 

\red{In all experiments, which we describe below, the same parameters have been used for the proposed method. Next we will indicate the values established for each of the parameters of the proposed method.}

\begin{figure}[!h]
\centering
\subfloat[][]{\includegraphics[width=.23\textwidth]{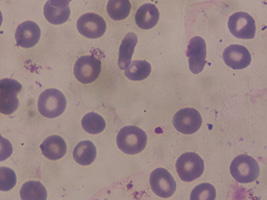}}\quad
			\subfloat[][]{\includegraphics[width=.23\textwidth]{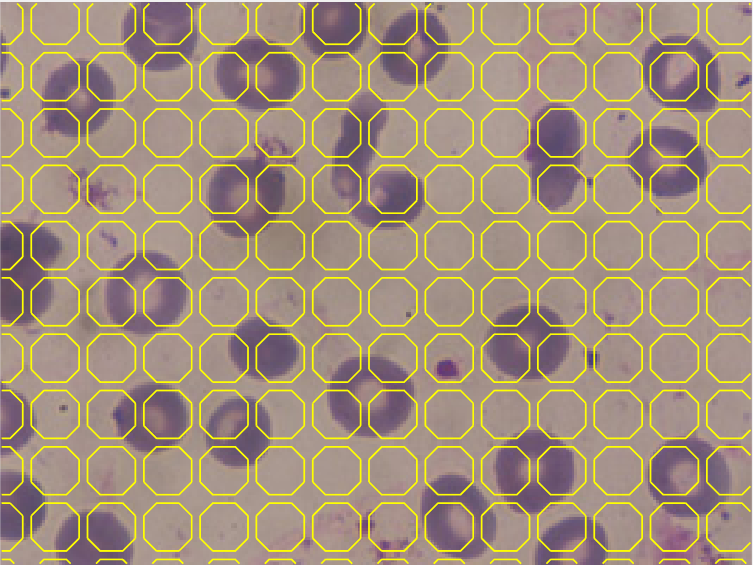}}\quad
			\subfloat[][]{\includegraphics[width=.23\textwidth]{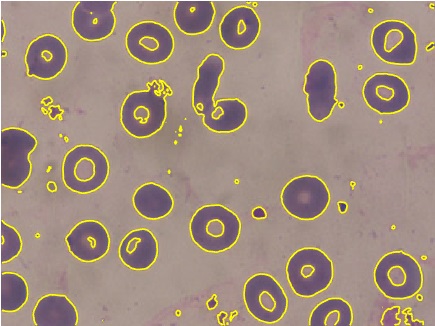}}\quad
			\subfloat[][]{\includegraphics[width=.23\textwidth]{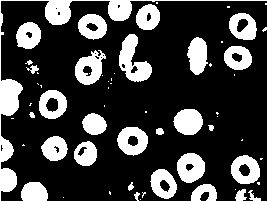}}
			\caption{\red{(a) Original input images. (b) Initial curve for the Chan-Vese active contour model. (c)  Evolution curve obtained for the Chan-Vese method for the image (a) and the initial condition (b). (d) Binary segmentation obtained from (c) and before small object elimination.}}
			\label{fig:chan-vese-init}
\end{figure}

\red{For the Chan-Vese segmentation method the regularization parameter $\mu$ is set to $0.2$, and the maximum number of iterations is set to 1000, but this value is nominal since the method converges much earlier for the images with we are working on. Different initial conditions will lead to various results especially in our multi-targets segmentation. So, our initial condition is fixed for all the experiments, and it is based on a hole net that is evolving over time using the Chan-Vese active contour model.  In figure \ref{fig:chan-vese-init} we show the initial condition,  and the obtained Chan-Vese segmentation for one of the images of our dataset.}

\red{As stated above in section \ref{ESFyCSF}, following \cite{asakura1996percentage} and \cite{fernandez2013estudio}  we set that we are detecting:
\begin{itemize}
\item An Elongated cell if ESF<=0.5. 
\item A Deformed slightly elongated cell if EFS > 0.5 and CSF  < 0.8.
\item A Normal shape if EFS > 0.5 and CSF  > 0.8.
\end{itemize}
}

\red{Finally, for the cluster detection and its analysis and classification of their cells we set the following parameters:
\begin{itemize}
\item In equation (\ref{Region}) we set $k=1.4$. This value is determined by systematic experimentation.  We did a 5 cross-validation over all the dataset, and we set the value with which we achieved the best performance. 
\item Deformed elongated cell if ESF<=0.6 (see \cite{gonzalez2015red}). 
\item Normal shape if EFS > 0.6 (see \cite{gonzalez2015red}).
\end{itemize}
}

\subsection{Experiment 1}\label{subsec:descr_exp1}	
The aim of this experiment was to validate the effectiveness of the proposed method for  the detection of the cells. For this purpose, the following performance index \cite{ruiz1999enfoque} was used.
	\begin{equation}\label{qualityF}
	\varPhi=\dfrac{\mathrm{DC}}{\mathrm{DC}+\mathrm{NDC}+\mathrm{NC}},  
	\end{equation}	
where
	\begin{itemize}
		\item DC is the quantity of detected cells. 
		\item NDC is the quantity of undetected cells.
		\item NC is the quantity of regions that does not belong to the detected cells.
	\end{itemize}	
The dataset was segmented using the Chan-Vese method to measure the effectiveness. 

\subsection{Experiment 2}\label{subsec:descr_exp2}	
	The process of classification using the descriptors (see Subsection \ref{ESFyCSF}) was applied to individual cells and overlapped cells. In the case of overlapping cells, before applying the aforementioned descriptors, an ellipse adjustment was performed (see Subsection \ref{elipse}). In order to check the performance of the classification, a $ 5\times1 $ cross-validation process for error estimation was used \cite{ferri1992comparison} and a confusion matrix with raw data and values of sensitivity, specificity,  precision, and F-measure for each class was obtained \cite{labatut2011accuracy,stapor2017evaluating}: 

	\begin{itemize}
	    \item Sensitivity ($R$) indicates the quantity of objects classified into the class to which they in fact belong, in relation to the total number of objects of that class.
	    \item Precision ($P$)  expresses the quantity of objects classified into the class to which they in fact  belong, in relation to the total number of objects identified as belonging to that class.
	    \item Specificity ($S$) expresses the quantity of objects classified as not belonging to one class to which they in fact  do not belong, in relation to the total number of objects that do not belong to that class. 
	   \item The F-measure ($F$) is  the harmonic mean of precision and sensitivity. Therefore, it is a symmetric function that gives the same relevance to both components. It can be interpreted as a measure of the overlapping between the true and estimated classes (other instances, i.e., true negative (TN), are ignored), ranging from 0 (no overlap at all) to 1 (complete overlap).
	\end{itemize}

\red{Let $\{n,e,o\}$ be the three true classes and the estimated classes be denoted by $\{\tilde{n},\tilde{e},\tilde{o}\}$. The  previous performance measures can be processed directly from the confusion matrix; see Table \ref{tab:conf_matrix}.  The term $n_{ij}$ ($1\leq i,j\leq 3$) corresponds to the number of cells assigned to  class $i$ by the classifier, whereas in fact they belong to class~$j$.}

\begin{table}[h]
    \centering
\begin{tabular}{l|cccc}
     & $\tilde{n}$ & $\tilde{e}$ & $\tilde{o}$  \\
     \hline
$n $& $n_{11}$ & $n_{12}$ & $n_{13}$ \\
$e $& $n_{21}$ & $n_{22}$ & $n_{23}$ \\
$o $& $n_{31}$ & $n_{32}$ & $n_{33}$ 
\end{tabular}
    \caption{\red{Confusion matrix of a three class classification problem.}}
    \label{tab:conf_matrix}
\end{table}

\red{Then for each class $i$, performance measures are computed as:}
\begin{align} \displaystyle
    R_i=\frac{n_{ii}}{\sum_{j=1}^3n_{ij}}, & \quad  P_i=\frac{n_{ii}}{\sum_{j=1}^3n_{ji}}, \\ 
    S_i=\frac{\sum\limits_{\substack{l=1 \\ l\not= i}}^3\sum\limits_{\substack{k=1 \\ k\not= i}}^3 n_{kl}}{\sum\limits_{\substack{l=1}}^3\sum\limits_{\substack{k=1 \\ k\not= i}}^3 n_{kl}}, &  \quad F_i=\frac{2R_i \cdot P_i}{R_i+ P_i}.
\end{align}

\subsection{Experiment 3}\label{subsec:descr_exp3}	
 In our selection of the methods with which to compare the method proposed in this paper, we considered various state-of-the-art automatic RBC automatic counter methods  based  on erythrocyte shape descriptors. The selected automatic counters are based  on very different theories, shape descriptors, and  frameworks covering a wide range and reflect the research efforts on this topic.

In \cite{acharya2018identification}, Acharya and Kumar presented a method for automatically classifying RBCs into two classes: normal cells (circular shape) and abnormal cells. After the image is converted to grey scale, image segmentation is performed using the global Otsu threshold method. Morphological operations and a modified watershed transform are applied to the image segmentation to separate the contiguous objects. Then, two possible means of counting the cells are possible: using the labeling algorithm and the CSF (form factor) to identify normal cells or using circular Hough transform. We considered these two possible means of counting, the first of which   we call  FF and the second HT.

Frejlichowski in \cite{frejlichowski2010preprocessing} presented a computer-assisted counter method based on a binary segmentation of the input image performed using modified thresholding based on fuzzy measures. Then, every RBC is localized and extracted. Using the template matching approach and shape description algorithm, every extracted erythrocyte is assigned to one of the three classes, normal, elongated, or other deformations. The shape descriptors are obtained using a normalized parametric polar transform (UNL-Transform) and then a 2D Fourier transform. We denote this method by UNL\textbf{-F}.   

A shape descriptor based on the principles of integral geometry was proposed in \cite{gual2015erythrocyte}. Using Crofton's formula, the authors proposed a new morphological erythrocyte shape descriptor of the RBC contour. The method operates on manually isolated cells in individual images of size $80\times 80$ pixels. Supervised classification algorithms can then be applied to them. No clusters are considered. The individual cells are segmented using an active contour-based segmentation method, and then, a k-NN classification algorithm is applied. We call this method CROFT.

The concept of generalized support functions as a representation of the shape properties of the isolated RBC contour was considered in \cite{gual2013shape}. This shape descriptor was used in the study reported in \cite{gual2015erythrocyte} for RBC classification. Again, this method operates only on isolated cells, without considering cells cluster.  We call this method SG.

These three methods operate only on isolated cells, without considering cluster cells. To apply them and to consider the shape descriptors established in each, we used as input for each algorithm the existing individual cells in the 30 analyzed images from our dataset. Therefore, the total number of cells classified was considerably lower. All the methods were implemented according to their descriptions in the original papers to which we referred. 

We denote by  2C the method proposed in this paper for a classification problem involving two classes (normal and others), and by  3C the method proposed in this paper for a classification problem involving three classes (normal, elongated, and other).

As indicated above, we present the results using a confusion matrix for each method, where the columns are the predicted class and the rows are the class to which the cells or clusters in fact belong. 

As for Experiment 2, we also present for each method the following standard measurements for each class, which can be obtained from the confusion matrices: precision, sensitivity, specificity, and F-measure (\cite{labatut2011accuracy,stapor2017evaluating}).

In our method, the universe of objects to be classified leads to imbalanced classes, because the overall field of view of the image is used, where the existing quantity of normal cells is considerably greater than that  of elongated or otherwise deformed cells. For this reason, the class balance accuracy (CBA) \cite{mosley2013} and Matthews correlation coefficient (MCC)  \cite{Gorodkin2004} measures were employed to evaluate the overall process. The study of performance measures for multiple class problems is an active research field~\cite{kautz2017generic,Branco2017}. These measures are defined as follows.
\begin{itemize}
    \item The CBA indicator, used in the study in \cite{Hauser2017,Branco2017} but originally defined in \cite{mosley2013},  is an alternative model performance indicator for use in the presence of class imbalance, because accuracy should be used only in situations where the overall performance is important and the class distributions are uniform. It is an overall accuracy measure built from an aggregation of individual class assessments. Individual accuracy assessments are calculated and then normalized by the number of extant classes. Information on the number of correctly predicted cases, contained in the diagonal elements, is normalized by either the total number of observations predicted to belong to the class or the actual number of observations in that class, according to which of the two numbers is larger.
    \item The definition of the MCC in a multiclass problem was originally presented in \cite{Gorodkin2004} (see also \cite{Jurman2012,Branco2017}). The MCC is a discrete version of Pearson's correlation coefficient and takes values in the interval $[-1, 1]$. The value is $1$ if there is a complete correlation, $0$ if there is no correlation, and $-1$ if there is a negative correlation.
\end{itemize}

For the two-class classification, we  used the counterpart measures.

To summarize, we compared the performance of three methods for a two-class classification problem (circular cells and other, the latter being composed of elongated cells and others) and of four methods for a three-class classification problem (circular, elongated, and other cells).  The three-class classification methods, UNL\textbf{-F}, SG, and CROFT, required the individual cells, whereas the 2C, HT,  FF, and 3C methods operated on the  full image. Finally, the  3C, 2C, and FF methods operated on clusters of cells.

\section{Results and Discussion}\label{section4}
\subsection{Experiment 1}
	
	Table \ref{tabla1}\footnote{The authors thank the authors of  \cite{fernandez2013estudio}, who executed their method to provide the data collected in Table 1, since the data collected in Table \ref{tabla1} of \cite{fernandez2013estudio} correspond with the initialization of the textures but not with the level set initialization.} shows  the results of the initial segmentation of regions belonging to erythrocytes. These results were compared with those in \cite{fernandez2013estudio}. The proposed method achieved a performance index of 96.46\% in the initial segmentation of the cells, showing an improvement on the results presented  in \cite{fernandez2013estudio}.
	The proposed method improves the  detection of the regions belonging to the cells and those not belonging to the cells.
	
	\begin{table}[h!] 
		\centering
		\begin{tabular}{lccccc}\toprule
			\textit{Methods}  & \textbf{DC} & \textbf{NDC} & \textbf{NC} & \textbf{$ \% $} \\  
			\hline 
			\ \textit{Our method}  &  436 & 6 & 10 & 96.46 \\ 
			\textit{\cite{fernandez2013estudio}} & 441 & 9 & 15 & 94.83 \\ 
			\hline 
		\end{tabular}
		\caption {Comparison of the results obtained by the proposed method and the method in \cite{fernandez2013estudio}, \red{where \textbf{DC} is the quantity of detected cells, \textbf{NDC} is the quantity of undetected cells, and \textbf{NC} is the quantity of regions that does not belong to the detected cells.}}\label{tabla1}
	\end{table}

The proposed method detected 26 (1 three-object cluster and 25 two-object clusters) from a total of 27 clusters, achieving an efficiency of 96.3\% in the detection of clusters. Although these results are good, they are not better than the results reported in  \cite{gonzalez2015red}, where an efficiency of 100\% was achieved.

The results of the classification into individual or cluster cells according to the experts'criteria using Equation (\ref{Region}) are shown in Table \ref{tabla2}. 
	
	\begin{table}[h!]
		\begin{center}			
			\begin{tabular}{lccccc}
				\toprule
				\textit{Type of regions} & \textbf{DC} & \textbf{NDC} & \textbf{NC}\\  
				\hline
				{\textit{single cells}}& 384& 5&6 \\ 
				{\textit{overlapped cells }}& 52 & 1& 4\\  
				\bottomrule
			\end{tabular}
			\caption{Detection of cells by type of region. \red{Recall that \textbf{DC} is the quantity of detected cells, \textbf{NDC} is the quantity of undetected cells, and \textbf{NC} is the quantity of regions that does not belong to the detected cells.}}\label{tabla2}.
		\end{center}
	\end{table}
	
A total of 384 individual cells and 52 cluster cells were detected, the latter belonging to 26 clustered regions. We remark that 18 of the 384 individual detected cells did not comply with the expert criteria for cluster detection. These cells were analyzed by means of applying ellipse adjustment to the cluster cells detected. For comparing the performance of our method with that of the SG, CROFT, and UNL\textbf{-F} methods, we employed only the cells that did comply with the expert criteria (366 single cells), because their analysis took into account the entire contour and not only the points of curvature.
	
\subsection{Experiment 2}
\begin{figure}[!ht]
\centering
\subfloat[][]{\includegraphics[width=.20\textwidth]{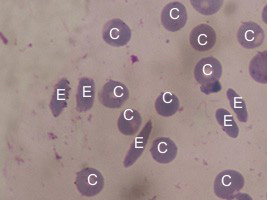}\label{fig:clasificacionEs}}\quad
			\subfloat[][]{\includegraphics[width=.20\textwidth]{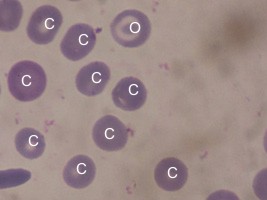}\label{fig:clasificacionEsp1}}\quad
			\subfloat[][]{\includegraphics[width=.20\textwidth]{img3.jpg}\label{fig:clasificacionSys}}\quad
			\subfloat[][]{\includegraphics[width=.20\textwidth]{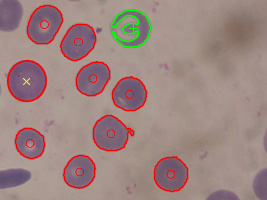}\label{fig:clasificacionSys1}}
			\caption{(a) and (b) Images classified by specialist. (c) and (d) Results of the classification produced by the proposed algorithm. \red{Circular cells in red, elliptical cells in blue and other cells in green.}}
			\label{fig:un_sub1}
\end{figure}

		Figures \ref{fig:clasificacionEs} and \ref{fig:clasificacionEsp1} show a sample of images classified by the specialist (C for circular/normal, E for elongated \red{and O for other}). Figures \ref{fig:clasificacionSys} and \ref{fig:clasificacionSys1} show the classification results of the proposed algorithm.  The cells are classified as normal (red boundary) or elongated (blue boundary). Cells classified as other deformations have a green boundary. The cells having areas that did not comply with Equation \ref{Region} were processed using an ellipse adjustment (see Section \ref{elipse}) and are identified by an ``X'' in the center, as is each cell in the cluster detected by the proposed algorithm.
		
	Table \ref{tabla3} shows the sensitivity values obtained for the classification of the isolated erythrocytes only. The values reported in  \cite{fernandez2013estudio} are in brackets. For circular RBCs, the results show a sensitivity value of 98\%, for elongated cells 92\%, and for other deformations 56\%. These values are higher than those reported in \cite{fernandez2013estudio}. When the normal and elongated cells are taken into account, the values obtained for the cells with other deformations do not change. No normal cells were classified as elongated. A percentage of normal cells was  classified as other deformations, although the percentage was smaller than that in the results presented in \cite{fernandez2013estudio}. In the case of elongated cells, a small percentage was  classified as normal (three cells, which represents 6\%) or as other deformations (one cell, which represents 2\%). The same occurs in the case of cells having other deformations that were  classified as normal and elongated: the latter value was higher than that obtained in the reference study. The results  can be affected by the sample preparation, which by its nature causes many cells to be damaged or broken, thus affecting their morphology.
	
	\begin{table}[h!]
		\begin{center}			
			\begin{tabular}{lcccc}
				\toprule
				\multirow{2}{*}{\textit{Class}}&\multicolumn{4}{c}{\textbf{\textit{CSF-ESF}}} \\ 
				\cmidrule{2-5} && \textbf{Normal} & \textbf{Elongated} & \textbf{Other Deform}\\  
				\cmidrule{1-5}				
				\textbf{Normal}	&& .98 (.87) & 0 (0) & .02 (.13) \\ 				
				\textbf{Elongated}&&.06 (.04)  &	.92 (.90) & .02 (.06) \\ 				
				\textbf{Other Deformations}&&.41 (.44)&	.04 (0) & .56 (.56)\\ 
				\bottomrule
			\end{tabular}
			\caption{Sensitivity values of the classification of isolated cells. \red{Recall that \textit{CSF} and \textit{ESF} are the circularity coefficient and the elliptical coefficient, respectively. }}\label{tabla3} 		
		\end{center}
	\end{table}

Table \ref{tabla4} shows the sensitivity, precision, specificity, and F-measure values of the classification of the isolated erythrocytes.  The sensitivity of the normal class was the highest (98.38\%), while that of the other deformations class was the lowest  (55.56\%). The highest precision and specificity values, 97.78\%, were obtained for the elongated class. For the other deformations class, a  specificity value of 97.48\% was exhibited; however, the precision value was lowest, 71.43\%. 
	\setlength\tabcolsep{2pt}
	\begin{table}[h!]
		\begin{center}
			
			\begin{tabular}{lccccc}
				\toprule
				\multirow{2}{*}{\textit{Class}}&\multicolumn{4}{c}{\textit{\textbf{CSF-ESF}}} \\ 
				\cmidrule{2-6} && \textbf{Sensitivity} & \textbf{Precision} & \textbf{Specificity}& \textbf{F-measure}\\  
				\cmidrule{1-6}
				
				\textbf{Normal}	&& 0.98 & 0.96 & 0.79 & 0.97\\ 
				
				\textbf{Elongated}&&0.92  &	0.98 & 0.95 & 0.95\\ 
				
				\textbf{Other Deformations}&&0.56&	0.71 & 0.97 & 0.63\\ 
				\bottomrule
			\end{tabular}
			\caption{Sensitivity, precision, specificity and F-measure values of the classification of isolated cells. \red{Recall that \textit{CSF} and \textit{ESF} are the circularity coefficient and the elliptical coefficient, respectively. }}\label{tabla4} 
			
		\end{center}
	\end{table}

Next, the clusters in the images were considered. Table \ref{tab3:solapamientos} shows the values obtained for the clusters after applying the ellipse adjustment. In this case, the cells were classified into only two classes: normal or elongated.
	
\setlength\tabcolsep{6pt}
\begin{table}[h]
	\begin{center}
		
		\begin{tabular}{lcccc}
			\toprule
			\multirow{2}{*}{\textit{Class}}&\multicolumn{4}{c}{\textit{\textbf{Ellipse adjustment}}} \\ 
			\cmidrule{2-5} && \textbf{Objects} & \textbf{Detected} & \textbf{Sensitivity}\\  
			\cmidrule{1-5}
			
			\textbf{Normal}	&& 38 & 37 & 0.97 \\ 
			
			\textbf{Elongated}&& 11  &10 & 0.91 \\ 
			\bottomrule
		\end{tabular}
		\caption{Classification of red blood cells, only in clusters.}\label{tab3:solapamientos} 
		
	\end{center}
\end{table}

The values presented in Table \ref{tab3:solapamientos} did not surpass those presented in \cite{gonzalez2015red} for either the normal or elongated classes. However, in the experiments in Gonzalez et al.'s study using real cells, a single region and not the entire image was analyzed. In the cluster, only one circular cell was classified as elongated and, vice versa, one elongated was classified as circular. These classification errors were due to the surrounding background of the clusters in which the cells were located. Moreover, three cells were classified by the specialist as belonging to the other deformations class. These three cells were classified as normal because of the specific characteristics of the system.

Table \ref{tab4:todo} shows the confusion matrix for the general classification, which includes the results for ellipse adjustment and the relationship between the CSF and ESF. Only one cell belonging to the other deformations class was classified as elongated, but these results do not affect the quality of the diagnosis. The  interesting result is that four elongated cells were classified as normal, which may lead to a misdiagnosis of the patient. These errors are related mostly to the threshold of the \textit{CSF} and \textit{ESF} and to a lesser extent to misclassification. 
	
	\begin{table}[h!]
		\begin{center}			
			\begin{tabular}{lcccc}
				\toprule
				\multirow{2}{*}{\textit{Class}}&\multicolumn{4}{c}{\textit{\textbf{TOTAL}}} \\ 
				\cmidrule{2-5} && \textbf{Normal} & \textbf{Elongated}& \textbf{Others deform.}\\  
				\cmidrule{1-5}				
				\textbf{Normal}	&& 342 & 1 & 5 \\ 				
				\textbf{Elongated} &&4  & 54 & 1 \\ 
				\textbf{Other deformations}	&&13  & 1   &15\\
				\bottomrule
			\end{tabular}
			\caption{Confusion matrix classification of red blood cells considering all cells detected.}\label{tab4:todo} 
		\end{center}
	\end{table}

Table \ref{tab4:todo_con_claib} shows  the sensitivity, precision, specificity, and F-Measure values for all regions of the images. The proposed method improved on the values previously obtained for the classification of each class, although some misclassifications occurred; for example, some elongated cells were classified as normal. The highest sensitivity and precision values were obtained for the normal and elongated classes, reaching 98.28\% and 96.43\%, respectively. 

\setlength\tabcolsep{2pt}
		\begin{table}[h!]
			\begin{center}
				\begin{tabular}{lccccc}
					\toprule
					\multirow{2}{*}{\textit{Class}}&\multicolumn{4}{c}{\textit{\textbf{TOTAL}}} \\ 
					\cmidrule{2-6} && \textbf{Sensitivity} & \textbf{Precision} & \textbf{Specificity} & \textbf{F-measure}\\  
					\cmidrule{1-6}					
					\textbf{Normal}	&&0.98& 0.95&0.81&0.97\\ 					
					\textbf{Elongated} &&0.92 & 0.96 & 0.99&0.94 \\ 
					\textbf{Other deformations}	&&0.52& 0.71   &0.99&0.60\\	
					\bottomrule
				\end{tabular}
				\caption{Classification of red blood cells considering isolated and cluster cells.}\label{tab4:todo_con_claib} 
			\end{center}
		\end{table}

\setlength\tabcolsep{6pt}

\begin{figure}[!t]
\centering
\begin{tabular}{ccc}
\subfloat[][]{\includegraphics[width=.25\textwidth]{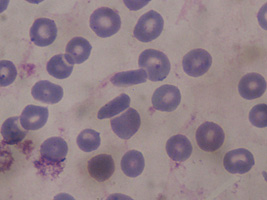}}  &  
\subfloat[][]{\includegraphics[width=.25\textwidth]{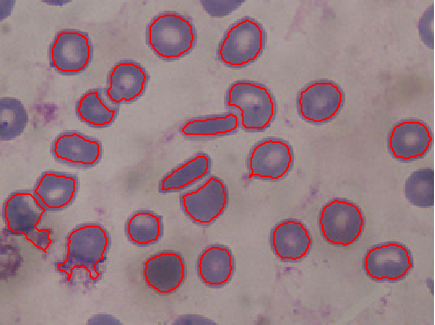}} & \subfloat[][]{\includegraphics[width=.25\textwidth]{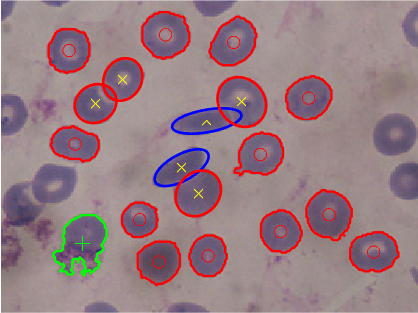}}\\
\subfloat[][]{\includegraphics[width=.25\textwidth]{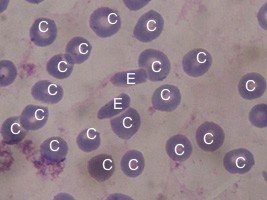}} &  
\subfloat[][]{\includegraphics[width=.25\textwidth]{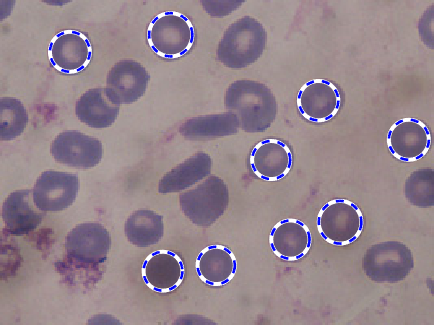}} 
\end{tabular}
			\caption{(a) Original image. (b) Cells classified as circular (normal cell)  by the method in \cite{acharya2018identification} using the FF algorithm. (c) Results of classification of the proposed algorithm (3C): in red, circular cells; in blue, elliptical cells; and in green, the other class. (d) Image cells classified by specialist.   (e) Cells classified as circular (normal cell)  by the method in \cite{acharya2018identification} using the HT algorithm.}
\label{fig:sub1}
\end{figure}

\subsection{Experiment 3}
\label{discussion-exp3}
When analyzing the results, we must take into account what type of errors can be more serious when these descriptors are applied in the clinical treatment of sickle cell disease. If normal cells are classified as  belonging to the elongated  or  other deformations class, as a result a greater  quantity of deformed cells will be reported, and it is possible that the specialist will decide that the patient's condition has deteriorated and that a more drastic treatment is required. This type of error is not very serious, because the treatment usually has no collateral effects. A more dangerous scenario would be one where deformed cells (elongated or other) are classified as normal. In this case, the specialist could decide that the patient is not at risk of a vaso-occlusive crisis, and the necessary treatment would not be administered.  Therefore, it is very desirable that any method that is applied to support diagnosis should guarantee a high specificity for normal cells (without producing a high percentage of false positives in this class),  while  maintaining  a high sensitivity for the other classes evaluated. In Figure~\ref{fig:sub1}, we show a comparison of the classification results of the proposed method and the results reported in~\cite{acharya2018identification}.

Table \ref{tab_disc1} shows the results for the 2C, HT, and FF classification methods and Table~\ref{tab_disc2} for the  3C, UNL\textbf{-F}, SG, and CROFT classification methods. \red{As a reminder, we used the same dataset for all the methods and we applied a 5 cross-validation. As we indicated in section~\ref{subsec:descr_exp3}, UNL\textbf{-F}, SG, and CROFT, require the individual cells, whereas the 2C, HT, FF, and 3C methods operate on the full image. Finally, the 3C, 2C, and FF methods operate on clusters of cells. The discrepancy in the number of cells is due to the existence of individual cells that did not meet the condition (\ref{Region}) and the existence of overlaps between the cells.}

\begin{table}
\begin{center}
\begin{tabular}{cc}	
			\begin{tabular}{lccc}
				\toprule
				\multirow{2}{*}{\textit{Class}}&\multicolumn{3}{c}{\textbf{\textit{HT \cite{acharya2018identification} }}} \\ 
				\cmidrule{2-4} && \textbf{Normal} & \textbf{Other}\\  
				\cmidrule{1-4}				
				\textbf{Normal}	&& 287 & 60 \\ 				
				\textbf{Other}&& 15  &	66  \\ 								
				\bottomrule
			\end{tabular}		
& 		
			\begin{tabular}{lccc}
				\toprule
				\multirow{2}{*}{\textit{Class}}&\multicolumn{3}{c}{\textbf{\textit{FF \cite{acharya2018identification} }}} \\ 
				\cmidrule{2-4} && \textbf{Normal} & \textbf{Other}\\  
				\cmidrule{1-4}				
				\textbf{Normal}	&& 352 & 2 \\ 				
				\textbf{Other}&& 76  &	11  \\ 								
				\bottomrule
			\end{tabular} \\ \\
			(a) & (b) \\ \\			
			\begin{tabular}{lccc}
				\toprule
				\multirow{2}{*}{\textit{Class}}&\multicolumn{3}{c}{\textbf{\textit{2C Proposed }}} \\ 
				\cmidrule{2-4} && \textbf{Normal} & \textbf{Other}\\  
				\cmidrule{1-4}				
				\textbf{Normal}	&& 342 & 6 \\ 				
				\textbf{Other}&& 17 &	71  \\ 								
				\bottomrule
			\end{tabular}
			\\ \\
			(c) & 
\end{tabular}
\end{center}
\caption{Confusion matrix for each one of the two-class classification methods: (a) HT method. (b) FF method. (c) Proposed method.} \label{tab_disc1}
\end{table}

\begin{table}
\begin{center}
\begin{tabular}{c}	
			\begin{tabular}{lcccc}
				\toprule
				\multirow{2}{*}{\textit{Class}}&\multicolumn{4}{c}{\textit{\textbf{UNL\textbf{-F} \cite{frejlichowski2010preprocessing}   }}} \\ 
				\cmidrule{2-5} && \textbf{Normal} & \textbf{Elongated}& \textbf{Others deform.}\\  
				\cmidrule{1-5}				
				\textbf{Normal}	&& 274 & 7 & 11 \\ 				
				\textbf{Elongated} && 6  & 40 & 1 \\ 
				\textbf{Other deformations}	&&  13  & 7   & 7\\
				\bottomrule
			\end{tabular}	\\ \\
(a)		
\\ \\
			\begin{tabular}{lcccc}
				\toprule
				\multirow{2}{*}{\textit{Class}}&\multicolumn{4}{c}{\textit{\textbf{ SG \cite{gual2015erythrocyte}  }}} \\ 
				\cmidrule{2-5} && \textbf{Normal} & \textbf{Elongated}& \textbf{Others deform.}\\  
				\cmidrule{1-5}				
				\textbf{Normal}	&& 277 & 3 & 12 \\ 				
				\textbf{Elongated} && 3  & 43 & 1 \\ 
				\textbf{Other deformations}	&&  10  & 2   & 15\\
				\bottomrule
			\end{tabular}	\\ \\
 (b) \\ \\			
			\begin{tabular}{lcccc}
				\toprule
				\multirow{2}{*}{\textit{Class}}&\multicolumn{4}{c}{\textit{\textbf{ CROFT \cite{gual2015erythrocyte}  }}} \\ 
				\cmidrule{2-5} && \textbf{Normal} & \textbf{Elongated}& \textbf{Others deform.}\\  
				\cmidrule{1-5}				
				\textbf{Normal}	&& 275 & 3 & 14 \\ 				
				\textbf{Elongated} && 8  & 38 & 1 \\ 
				\textbf{Other deformations}	&&  11  & 2   & 14\\
				\bottomrule
			\end{tabular}	\\ \\
 (c) \\ \\			
			\begin{tabular}{lcccc}
				\toprule
				\multirow{2}{*}{\textit{Class}}&\multicolumn{4}{c}{\textit{\textbf{ 3C Proposed  }}} \\ 
				\cmidrule{2-5} && \textbf{Normal} & \textbf{Elongated}& \textbf{Others deform.}\\  
				\cmidrule{1-5}				
				\textbf{Normal}	&& 342 & 1 & 5 \\ 				
				\textbf{Elongated} && 4  & 54 & 1 \\ 
				\textbf{Other deformations}	&&  13  & 1   & 15\\
				\bottomrule
			\end{tabular}	\\ \\
 (d)  
\end{tabular}
\end{center}
\caption{Confusion matrix for the three-class classification methods. (a)  UNL\textbf{-F} \cite{frejlichowski2010preprocessing}. (b)  SG \cite{gual2015erythrocyte}. (c) CROFT \cite{gual2015erythrocyte}. (d) 3C proposed method.} \label{tab_disc2} 
\end{table}

\begin{table}
\begin{center}
\begin{tabular}{c}	
				\begin{tabular}{lccccc}
					\toprule
					\multirow{2}{*}{\textit{Class}}&\multicolumn{4}{c}{\textit{\textit{HT \cite{acharya2018identification} }}} \\ 
					\cmidrule{2-6} && \textbf{Precision} & \textbf{Sensitivity} & \textbf{ Specificity} & \textbf{F-measure}\\  
					\cmidrule{1-6}					
					\textbf{Normal}	&&   0.95 &	0.83	& 0,81 &	0,88 \\ 					
					\textbf{Other } &&   0.52 &	0.81	& 0.83 & 	0.64  \\ 
					\bottomrule
				\end{tabular}	\\ \\
(a)		
\\ \\
				\begin{tabular}{lccccc}
					\toprule
					\multirow{2}{*}{\textit{Class}}&\multicolumn{4}{c}{\textit{\textit{FF \cite{acharya2018identification} }}} \\ 
					\cmidrule{2-6} && \textbf{Precision} & \textbf{Sensitivity} & \textbf{ Specificity} & \textbf{F-measure}\\  
					\cmidrule{1-6}					
					\textbf{Normal}	&&  0.82	& 0.99 &	0.13	& 0.90 \\ 					
					\textbf{Other} && 0.85 & 0.13	& 0.99	& 0.22  \\ 
					\bottomrule
				\end{tabular}	\\ \\
 (b) \\ \\			
				\begin{tabular}{lccccc}
					\toprule
					\multirow{2}{*}{\textit{Class}}&\multicolumn{4}{c}{\textit{\textit{2C Proposed }}} \\ 
					\cmidrule{2-6} && \textbf{Precision} & \textbf{Sensitivity} & \textbf{ Specificity} & \textbf{F-measure}\\  
					\cmidrule{1-6}					
					\textbf{Normal}	&&  0.95	& 0.98 &	0.81	& 0.97 \\ 					
					\textbf{Other } &&   0.92 &	0.81 &	0.98 &	0.86  \\ 
					\bottomrule
				\end{tabular}	\\ \\
 (c) 	
\end{tabular}
\end{center}
\caption{Performance measures for each one of the two-class classification methods. (a) HT method. (b) FF method. (c) Proposed method.} \label{tab_disc3}
\end{table}

\setlength\tabcolsep{2pt}
\begin{table}[h]
\begin{center}
\begin{tabular}{c}	
				\begin{tabular}{lccccc}
					\toprule
					\multirow{2}{*}{\textit{Class}}&\multicolumn{4}{c}{\textit{\textit{\textbf{UNL\textbf{-F} \cite{frejlichowski2010preprocessing}   } }}} \\ 
					\cmidrule{2-6} && \textbf{Precision} & \textbf{Sensitivity} & \textbf{ Specificity} & \textbf{F-measure}\\  
					\cmidrule{1-6}					
					\textbf{Normal}	&&   0.94 &	0.94	& 0.74	& 0.94 \\ 					
					\textbf{Elongated} &&   0.74	& 0.85	& 0.96 &	0.79  \\ 
					\textbf{Other deformations}	&&  0.37 &	0.26	& 0.96 & 0.30 \\	
					\bottomrule
				\end{tabular}	\\ \\
(a)		
\\ \\
				\begin{tabular}{lccccc}
					\toprule
					\multirow{2}{*}{\textit{Class}}&\multicolumn{4}{c}{\textit{\textbf{ SG \cite{gual2015erythrocyte}  }}} \\ 
					\cmidrule{2-6} && \textbf{Precision} & \textbf{Sensitivity} & \textbf{ Specificity} & \textbf{F-measure}\\  
					\cmidrule{1-6}					
					\textbf{Normal}	&&  0.96	& 0.95	& 0.82	& 0.95 \\ 					
					\textbf{Elongated} && 0.89	& 0.91	& 0.98& 	0.91  \\
					\textbf{Other deformations}	&&  0.54 &	0.56	& 0.96	& 0.54 \\
					\bottomrule
				\end{tabular}	\\ \\
 (b) \\ \\			
				\begin{tabular}{lccccc}
					\toprule
					\multirow{2}{*}{\textit{Class}}&\multicolumn{4}{c}{\textit{\textbf{ CROFT \cite{gual2015erythrocyte} }}} \\ 
					\cmidrule{2-6} && \textbf{Precision} & \textbf{Sensitivity} & \textbf{ Specificity} & \textbf{F-measure}\\  
					\cmidrule{1-6}					
					\textbf{Normal}	&&  0.94	& 0.94	& 0.74& 0.94 \\ 					
					\textbf{Elongated} &&   0.88 & 	0.81 &	0.98 &	0.84   \\
					\textbf{Other deformations}	&&  0.48	& 0.52	& 0.96	& 0.5 \\
					\bottomrule
				\end{tabular}	\\ \\
 (c) \\ \\	
 \begin{tabular}{lccccc}
					\toprule
					\multirow{2}{*}{\textit{Class}}&\multicolumn{4}{c}{\textit{\textbf{ 3C Proposed  }}} \\ 
					\cmidrule{2-6} && \textbf{Precision} & \textbf{Sensitivity} & \textbf{ Specificity} & \textbf{F-measure}\\  
					\cmidrule{1-6}					
					\textbf{Normal}	&&  0.96	& 0.98 & 	0.81 &	0.97 \\ 					
					\textbf{Elongated} &&   0.98	& 0.92	& 0.997	& 0.95  \\
					\textbf{Other deformations}	&& 0.71 &	0.56	& 0.98 &	0.63 \\
					\bottomrule
				\end{tabular}	\\ \\
 (d) 
\end{tabular}
\end{center}
\caption{Performance measures for each one of the three-class classification methods. (a) UNL\textbf{-F} method. (b) SG method. (c) CROFT method. (d) Proposed method.} \label{tab_disc4}
\end{table}

Our objective was to design a method that can be applied to support the diagnosis of patients with sickle cell disease. To achieve this, our method should not show a tendency to classify elongated or cells with other deformations as circular, because,  as mentioned previously,  a specialist could conclude that the patient's condition has improved and he/she is not at risk of a vaso-occlusive crisis. Specialists consider cells  with other deformations as elongated, because it is not possible to be sure whether  these  cells are  deformed  circular or elongated cells, and, because the treatment does not have collateral effects, the difference is not crucial.

Conversely, if circular cells are misclassified, specialists could conclude that the patient's condition has deteriorated and they could prescribe a more drastic treatment. As mentioned previously, this type of error would not have very negative consequences, because the treatment has no side effects. 

Therefore, a good method for this diagnostic application must offer high sensitivity for elongated cells and cells with other deformations to minimize the classification of these two classes of cells as circular cells, thus  minimizing the number of false negatives and avoiding the diagnosis of a false improvement in the patient's condition. It also must offer a high specificity in the classification of circular cells to avoid their classification as elongated or as cells with other deformations. 

Tables \ref{tab_disc3} and  \ref{tab_disc4} show that our method obtained a high sensitivity value for elongated/other cells and a high specificity for circular cells in the two-class classification problem.    In the three-class classification problem, our method obtained the best sensitivity for elongated cells, a high sensitivity for  cells with other deformations, and a high specificity for circular cells. In addition, our method obtained the best F-measure values for all classes, which means that, in general, our method achieves the best minimization of false negatives for elongated/other cells and false positives for circular cells.

To measure the accuracy of the diagnoses of patients with sickle cell disease, taking into account the  minimization of  false negatives for elongated/other cells and false positives for circular cells mentioned above, we defined a score called the sickle cell disease diagnosis support score ({\sl SDS-score}) as the quotient of the sum of true positives of the three classes and the number of elongated cells classified as cells with other deformations and vice versa divided by the sum of the previous numerator and the sum of the incorrect classifications related to  circular cells. The SDS-score is intended to constitute an indicator that the results provided by the method are useful for the diagnosis support of the considered disease. In the case of sickle cell disease, it coincides with the accuracy of the associated two-class problem. The advantage of our interpretation is that this measure could be generalized to other diseases. \red{The  {\sl SDS-score} is processed directly from the confusion matrix (see Table \ref{tab:conf_matrix}) and is defined as:}

%\setlength\tabcolsep{2pt}
\iffalse
\begin{table}[h]
    \centering
\begin{tabular}{l|cccc}
     & $\tilde{n}$ & $\tilde{e}$ & $\tilde{o}$  \\
     \hline
$n $& $n_{11}$ & $n_{12}$ & $n_{13}$ \\
$e $& $n_{21}$ & $n_{22}$ & $n_{23}$ \\
$o $& $n_{31}$ & $n_{32}$ & $n_{33}$ 
\end{tabular}
    \caption{Confusion matrix of a three class classification problem.}
    \label{tab:conf_matrix}
\end{table}

\textst{Let $\{n,e,o\}$ be the three true classes and the estimated classes be denoted by  $\{\tilde{n},\tilde{e},\tilde{o}\}$. The  {\sl SDS-score} is processed directly from the confusion matrix; see Table \ref{tab:conf_matrix}.  The term $n_{ij}$ ($1\leq i,j\leq 3$) corresponds to the number of cells assigned to  class $i$ by the classifier, whereas in fact they belong to class~$j$. The {\sl SDS-score} is defined as}
\fi
\[
\mbox{\sl{SDS-score}} = \frac{\sum\limits_{i=1}^3 n_{ii} + n_{23}+n_{32}}{\sum\limits_{i=1}^3\sum\limits_{j=1}^3n_{ij}}. 
\]

For the sake of clarity, the performance indicators CBA and MCC can be computed directly from the confusion matrix following the methods in \cite{mosley2013} and \cite{Jurman2012}, respectively: 
\begin{align*}
   & CBA=\frac{1}{3}\sum_{i=1}^3 \frac{n_{ii}}{\max\left( \sum\limits_{j=1}^3n_{ij},\sum\limits_{j=1}^3n_{ji}\right)}, \\
    & MCC=\frac{\sum\limits_{k,l,m=1}^3\left( n_{kk}n_{ml} - n_{lk}n_{km} \right)}{\sqrt{\sum\limits_{k=1}^3\left[ \left( \sum\limits_{l=1}^3 n_{lk}\right)  \left( \sum\limits_{f,g=1 \atop f\not= k}^3  n_{gf} \right)\right]} \sqrt{\sum\limits_{k=1}^3\left[ \left( \sum\limits_{l=1}^3 n_{kl}\right)  \left( \sum\limits_{f,g=1 \atop f\not= k}^3  n_{fg} \right)\right]
}}.
\end{align*}

Table \ref{tab_disc5} shows that, in this study, our method achieved the  best {overall performance} in the two- and three-class classification problems. Therefore, we can conclude that our method is the best in terms of supporting the diagnosis of patients with sickle cell disease.  

\begin{figure}[!ht]
\centering
\subfloat[][]{\includegraphics[width=.20\textwidth]{HCSS015C3.jpg}\label{fig:clasificacionEs1}}\quad
			\subfloat[][]{\includegraphics[width=.20\textwidth]{img3.jpg}\label{fig:clasificacionmetpro}}\quad
			\subfloat[][]{\includegraphics[width=.20\textwidth]{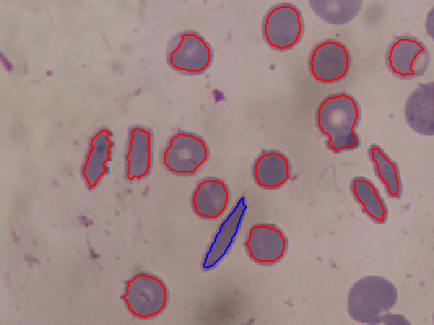}\label{fig:clasificacionFF}}\quad
			\subfloat[][]{\includegraphics[width=.20\textwidth]{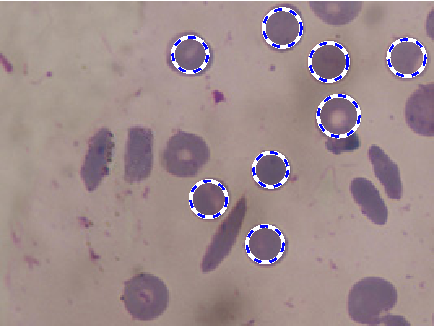}\label{fig:clasificacionHT}}
			\caption{\red{(a)  Image classified by specialist. (b) Result of the classification produced by the proposed algorithm. Circular cells in red, elliptical cells in blue. (c) Cells classified as circular (in red) and elliptical (in blue) by the method in \cite{acharya2018identification} using the FF algorithm. (d) Cells classified as circular (in dashed blue) by the method in \cite{acharya2018identification} using the HT algorithm. }}
			\label{fig:comp2}
\end{figure}

\red{In addition to figure \ref{fig:sub1}, we show in figure \ref{fig:comp2} another example in which the performance of the proposed method outperform the methods with  which we compare in this work that work with a full-range image.}

We would like to remark that when applied to the three-class classification problem our method yielded better results in the other measures (see Tables~\ref{tab_disc3} and \ref{tab_disc4}) than when applied to the two-class classification problem. However, specialists consider cells with other deformations as elongated for the purpose of prescribing medication. For this reason, this cell class is considered a rejection class.   

\begin{table}[h]
\begin{center}
\begin{tabular}{ccc}	
			\begin{tabular}{lccc}
				\toprule
				\textbf{2-Class} & \textbf{\sl{SDS-score}}& \textbf{CBA}& \textbf{MCC}\\  
				\cmidrule{1-4}				
				  HT & 0.82& 0.82 &0.55 \\ 				
				  FF  &	0.82&0.56 &0.28 \\ 	
				  2C & 0.95& 0.89&0.84 \\ 		
				\bottomrule
			\end{tabular}	& & 	\begin{tabular}{lccc}
				\toprule
				\textbf{3-Class} & \textbf{\sl{SDS-score}}& \textbf{CBA}& \textbf{MCC}\\  
				\cmidrule{1-4}				
				  UNL\textbf{-F} & 0.90 &0.65 &0.64\\	
				  SG  &	0.92 & 0.79&0.75 \\ 	
				  CROFT & 0.90 & 0.74&0.69\\ 	
				  3C & 0.95 & 0.80& 0.82 \\			
				\bottomrule
			\end{tabular} \\
(a)	&	& (b) 
\end{tabular}
\end{center}
\caption{Sickle cell disease diagnosis support score, class balance accuracy, and Matthews correlation coefficient for all  compared methods.} \label{tab_disc5}
\end{table}

Our method is competitive with the state-of-the art methods, because it achieves high performance measurements. We would like to remark again that our method achieved the {best F-measure value for all classes and the best SDS-score, CBA, and MCC values.} Moreover, our method is able to handle overlapped clusters of cells. \red{Besides, the most important limitation in classifying erythrocytes shapes is related to the way in which laboratories prepare blood samples for microscopy analysis. They usually use the dragging technique that can generate shadows, artifacts, deformations of the erythrocytes and the overlapping of the cells in the captured image.}

The dataset used in the experiments is an imbalanced dataset, because  samples from sickle cell disease patients  include more circular cells than cells of other shapes.

\section{Conclusions}

The morphological analysis of digital image structures is essential in the process of extracting features of interest from the analyzed edges  for subsequent use in appropriate classification algorithms. In this paper, we presented a method for classifying erythrocytes in peripheral blood samples as normal cells, elongated cells, and those with other deformations using elemental coefficients, the \textit{CSF} and \textit{ESF}. The proposed method has  an advantage in the analysis of partially occluded cells in clusters: an elliptic adjustment is conducted to allow the analysis of erythrocytes with discoidal and elongated shapes.  Furthermore,  the proposed method does not require an input image preprocessing step  to remove noise, as is usual in the state-of-the-art methods.

The regions of interest were obtained by using the Chan-Vese segmentation algorithm, leading to an improvement on the results reported in~\cite{gonzalez2015red}. The results obtained in the classification of single cells show that the performance of the proposed method is better than that of the method presented in~\cite{fernandez2013estudio} for normal and elongated cells. In the analysis of cell clusters, the proposed method operates on the full range of images and achieves a high  efficiency level,  while the method in \cite{gonzalez2015red} operates only on isolated cells.  However, it is valid to mention that some clusters are  formed by unloading  cells that belong to the class with other deformations, despite  the noise inherent in the images.

The morphological classification of erythrocytes is very important in the treatment of  sickle cell disease patients. For the clinical treatment and diagnostic support of this disease,  a high F-measure and high values of the corresponding multiclass and two-class overall performance measures are desirable. The results achieved by the proposed method are suitable for this purpose. In comparison with  the methods in \cite{acharya2018identification,fernandez2013estudio,frejlichowski2010pre,gual2015erythrocyte}, our method achieved a superior performance, obtaining for the two- and three-class classification the highest F-measure value and the highest values for several overall multiclass performance measures.

We include in the results the confusion matrices with the raw data to allow researchers to more easily compute other metrics. The  dataset we used is available at \url{http://erythrocytesidb.uib.es/}. For the sake of scientific progress, it would be beneficial if authors published their raw data and the image datasets that they used.

\begin{acknowledgements}
We acknowledge the support of the Agencia Estatal de Investigaci{\'o}n (AEI) and the European Regional Development Funds (ERDF)  of  Projects TIN2016-81143-R (AEI/FEDER, UE) and TIN2016-75404-P, and the support of the Government of the Balearic Islands and the University of the Balearic Islands of projects OCDS-CUD2016/01 and OCDS-CUD2017/05. We also thank the Mathematics and Computer Science Department at the University of the Balearic Islands for its support.
\end{acknowledgements}

%\section*{References}

\bibliographystyle{spmpsci}      % mathematics and physical sciences

\bibliography{mybibfile}   % name your BibTeX data base

\section*{Authors' biography}

\textbf{Wilkie Delgado-Font}: He received the degree in computers engineer in 2012 from the Universidad de Oriente, Cuba. Currently, he is finishing the studies for the Master degree in Computer Science, by the Universidad de Oriente and developing the PhD thesis. He is an assistant professor in the Computing Department of the Faculty of Natural and Exact Sciences of the Universidad de Oriente since 2012. He belongs to the Image Processing and Computer Visioning Group of the Computer Science Department conducting research in the field of quantitative morphological analysis of samples biological. He is a member of the Sociedad Cubana de Matem\'atica Computaci\'on since 2018. Research interests include topics of medical image processing, bioinformatics and application development. The research that he has developed in this field has earned him the presentation in national and international events, as well as the publication of the results obtained in specialized journals.

\textbf{Miriela Escobedo-Nicot}:She received a degree in Computer Science in 2007, a Master's Degree in Computer Science in 2009 and a PhD in Technical Sciences in 2018, all from the Universidad de Oriente, Cuba. From 2007 she is a professor in the Computer Science Department of the Faculty of Natural and Exact Sciences of the Universidad de Oriente. She is member of the Image Processing and Computer Vision Group of the Computer Science Department and the Sociedad Cubana de Matem\'atica Computaci\'on since 2007. Research interests include topics of pattern recognition, bioinformatics and medical image processing acquired through different techniques, optics such as digital holographic microscopy and light field microscopy. The academic production includes articles in specialized magazines, works in events, as well as cooperation in specialized books. The current research is oriented to the automatic morphological quantification of biological processes developed in experimental environments, using digital image processing techniques.

\textbf{Manuel Gonz\'alez-Hidalgo}: He was born in the province of Leon in 1964, Spain. Degree in Mathematics, Specialty General Mathematics, Guidance Mathematical Analysis, by the University of Valencia in 1988. Ph.D. in Computer Science from the University of the Balearic Islands (UIB) in 1995. He is currently an Associate Professor with the Department of Mathematics and Computer Science of the UIB. The research areas of interest included computer vision, image analysis, modeling and animation of deformable objects, analysis and synthesis of human movement, medical imaging and 3D modeling, and recently the study of aggregation operators and their applications to image processing and analysis, focusing on the fuzzy mathematical morphology and its applications. Currently he is working in Soft Computing techniques and its applications to biomedical image analysis. Member of the research group SCOPIA 'Soft Computing, Image Processing and Aggregation' and regular collaborator with the research group 'Computer Graphics and Vision and AI Group (UGiVIA)'. This research activity has been reflected in scientific publications in international journals and in papers presented at national and international conferences. It is referee for several journals of reference in his research fields, has organized several congress and special sessions, and different scientific activities.

\textbf{Silena Herold-Garcia}: She obtained her PhD in Computer Science by the Universidad de Oriente. She is a professor in the Computer Science Department of the Faculty of Natural and Exact Sciences of the Universidad de Oriente. She is member of the Image Processing and Computer Vision Group of the Computer Science Department and the Sociedad Cubana de Matem\'atica Computaci\'on since 2007. Research interests include topics of pattern recognition, bioinformatics and medical image processing acquired through different techniques, optics such as digital holographic microscopy and light field microscopy.

\textbf{Antoni Jaume-i-Cap\'o}: European PhD in Computer Science by the University of the Balearic Islands (UIB). He is associate professor in the Department of Mathematics (DMI) and Computer Science at UIB. His lines of research are computer vision, vision-based interfaces, interactive systems for motor rehabilitation, medical imaging processing and human-based computing for image processing. He is researcher at the Computer Graphics and Vision and Artificial Intelligence Group (UGiVIA), a group that is part of the University Institute for Research in Health Sciences (IUNICS). 

\textbf{Arnau Mir}: In the year 1987, the professor Arnau Mir finished his bachelor's degree. From 1987 until the year 1994, he was working in the group of dynamic systems of the University of Barcelona. Their PhD was the result of this collaboration. Since the year 1995 until the year 2006, he was working of one hand with professor Pedro Salvador, professor of CSIC in the analysis of photoelectrochemical images for the characterization of semiconductors and on the other hand he was a member of the group of Graphic and Vision of the University of the Balearic Islands (UIB) and he worked in the research area of deformation and simulation of objects. From the year 2006, he is a member of the research group of Bioinformatics  and Computational Biology (BIOCOM) of the UIB. Specifically, he is working in the balance indexes of rooted and non rooted binary and multifurcating trees. Also, he is a collaborator in the research group of Soft Computing and Image Processing and Aggregation (SCOPIA). His work is focused in the boundary detection and noise removal of images (in particular, medical images) using fuzzy techniques

\end{document}